\documentclass[sigconf,screen]{acmart}

\copyrightyear{2026}
\acmYear{2026}
\setcopyright{cc}
\setcctype{by-nc-nd}
\acmConference[FSE Companion '26]{34th ACM Joint European Software Engineering Conference and Symposium on the Foundations of Software Engineering}{July 05--09, 2026}{Montreal, QC, Canada}
\acmBooktitle{34th ACM Joint European Software Engineering Conference and Symposium on the Foundations of Software Engineering (FSE Companion '26), July 05--09, 2026, Montreal, QC, Canada}
\acmDOI{10.1145/3803437.3806712}
\acmISBN{979-8-4007-2636-1/2026/07}

\PassOptionsToPackage{prologue,dvipsnames}{xcolor}

\usepackage[utf8]{inputenc}
\usepackage{ifthen}
\usepackage{hyperref}

\usepackage[nomessages]{fp}%
\usepackage[prologue,dvipsnames]{xcolor}
\usepackage{soul}
\usepackage{enumitem}
\usepackage{subcaption}
\usepackage{multirow}
\usepackage{fancybox}
\usepackage{framed}

\newcommand{\secref}[1]{Sec.~\ref{#1}}
\newcommand{\figref}[1]{Fig.~\ref{#1}}
\newcommand{\tabref}[1]{Tab.~\ref{#1}}

\newcommand\xofy[3]{%
    \FPeval{perc}{round(100.0 / #2 * #1, 1)}%
    \ifthenelse{\equal{#3}{}}{%
        {#1} of {#2} (\perc\%)%
    }%
    {%
        {#1} of {#2} {#3} (\perc\%)%
    }%
}

\newcommand\xofyp[2]{%
    \FPeval{perc}{round(100.0 / #2 * #1, 1)}%
    {#1} of {#2} -- \perc\%%
}

\newcommand\perc[2]{%
    \FPeval{perc}{round(100.0 / #2 * #1, 1)}%
    \perc%
}

\cornersize*{5pt} 
\newenvironment{cframe}[1]{%
  \begin{framed}
    \centering
    \vspace{-1.2em}
    \colorbox{white}{\space\textbf{#1}\space}
    \vspace{-0em}
    \par
    \list{}{\leftmargin=7pt \rightmargin=7pt}\item\relax
    \sloppy
}{%
  \endlist%
  \end{framed}
}

\newenvironment{siderules}{%
  \setlength{\FrameRule}{1.5pt}%
  \setlength{\fboxsep}{0pt}
  \MakeFramed{\advance\hsize-\width\FrameRestore}%
  \list{}{\leftmargin=2pt \rightmargin=0pt}\item\relax
}{%
  \endlist%
  \endMakeFramed
}

\newcommand{\keyobservation}[1]{%
    \begin{siderules}
    \noindent\textit{\textbf{Observation:} #1}
    \end{siderules}
    
    \phantom{}\\[-1.75em]
}

\definecolor{darkblue}{rgb}{0.0,0.0,0.55}
\definecolor{cornflowerblue}{rgb}{0.39, 0.58, 0.93}
\definecolor{silver}{rgb}{0.75, 0.75, 0.75}
\definecolor{lightcoral}{RGB}{240,128,128}
\definecolor{firebrick}{rgb}{0.698,0.133,0.133}

\begin{document}

\title{Trust the AI, Doubt Yourself: The Effect of Urgency on Self-Confidence in Human-AI Interaction}

\author{Baran Shajari}
\orcid{0009-0005-3133-6441}
\affiliation{
  \institution{McMaster University}
  \city{Hamilton}
  \country{Canada}
}
\email{shajarib@mcmaster.ca}

\author{Xiaoran Liu}
\orcid{0009-0000-9908-7406}
\affiliation{
  \institution{McMaster University}
  \city{Hamilton}
  \country{Canada}
}
\email{liu2706@mcmaster.ca}

\author{Kyanna Dagenais}
\orcid{0009-0007-6304-3971}
\affiliation{
  \institution{McMaster University}
  \city{Hamilton}
  \country{Canada}
}
\email{dagenaik@mcmaster.ca}

\author{Istvan David}
\orcid{0000-0002-4870-8433}
\affiliation{
  \institution{McMaster University}
  \city{Hamilton}
  \country{Canada}
}
\email{istvan.david@mcmaster.ca}

\renewcommand{\shortauthors}{Shajari et al.}

\begin{abstract}
Studies show that interactions with an AI system fosters trust in human users towards AI. An often overlooked element of such interaction dynamics is the (sense of) urgency when the human user is prompted by an AI agent, e.g., for advice or guidance. In this paper, we show that although the presence of urgency in human-AI interactions does not affect the trust in AI, it may be detrimental to the human user's self-confidence and self-efficacy. In the long run, the loss of confidence may lead to performance loss, suboptimal decisions, human errors, and ultimately, unsustainable AI systems.
Our evidence comes from an experiment with 30 human participants.
Our results indicate that users may feel more confident in their work when they are eased into the human-AI setup rather than exposed to it without preparation. We elaborate on the implications of this finding for software engineers and decision-makers.
\end{abstract}



\maketitle

\section{Introduction}

Recent trends in AI systems have shifted their \textit{modus operandi} from simple stimulus-response systems to more autonomous ones~\cite{mseer2025autonomous}. Mixed-initiative~\cite{schiaffino2004user} and agentic AI~\cite{nisa2025agentic} systems, specifically, are autonomous AI systems equipped with the ability to proactively prompt the human, e.g., for guidance, advice, or approval~\cite{nisa2025agentic}. While evidently of high utility~\cite{benchaaben2025utility}, these classes of AI systems introduce novel challenges in human-AI interaction, one of which, as our study shows, is the reduced self-confidence of humans.

Studies that investigate the relationship of human and AI collaborators often focus on trust~\cite{kumar2025establishing} and its elements, such as the predictability~\cite{nordheim2019initial} and explainability of the AI agent~\cite{afroogh2024trust}. Evidently, such properties tend to improve as a result of extended interactions between humans and AI---such as human-AI collaboration~\cite{wan2024it}, guidance~\cite{wu2023human-in-the-loop}, cooperation~\cite{zhang2023trust} and joint work~\cite{denno2024cognitive}. As we show, humans' confidence does not necessarily coincide with increased trust. Moreover, \textbf{it is possible that trust towards AI increases while the user's confidence (in their own work and role) deteriorates} due to unprepared urgency.
We hypothesize that this cognitive asymmetry may lead to the elevated anxiety in users that has been reported in numerous studies~\cite{sowislo2013does}\cite{juth2008how}\cite{diebel2025when}.

To guide our investigation, we formulate two research questions.
\begin{description}[leftmargin=0.4cm, labelwidth=!, itemindent=0pt]
    \item[RQ1.] \textbf{Can we corroborate that interaction with the AI agent improves user trust in our experiment?}\label{RQ1}\\
    First, we aim to establish consistency with the state of the art by demonstrating that in our setup, trust in the AI agent improves in response to human-AI interactions.
    
    \item[RQ2.] \textbf{How does perceived urgency in human-AI interaction affect human users' relationship to the AI agent?}\label{RQ2}\\
    We aim to understand whether and which human attitudes change when subjecting human users to time-pressured interactions with an AI agent.
\end{description}

To answer these research questions, we conducted an experiment with 30 human participants, in which we tasked them with solving an interactive assignment in collaboration with an AI agent.\footnote{Replication package: \url{https://doi.org/10.5281/zenodo.19362930}.} By assessing the participants' trust attitudes before and after the experimental task, we corroborate that trust improves in mixed-initiative AI settings (RQ1)---a positive effect; but we observe that humans react to time pressure with lowered confidence in one's own work (RQ2)---a negative effect.

Our results suggest that AI systems with the ability to prompt humans may have a negative impact on the long-term sustainability of human-AI joint work without careful and gradual introduction of AI tools, and without proper upskilling, education, and training. These implications are important for vendors of AI-based solutions and organizational leaders alike, as they draw the attention to the need for a coordinated effort when deploying such systems. As well, software engineers as they are responsible for developing urgency cues into system behavior~\cite{abrahao2025software}.

\section{Background and Related Work}\label{sec:background}

\subsection{Trust in AI}\label{sec:trust-in-ai}
 
One of the common definitions of trust is ``\textit{the willingness of a party to be vulnerable to the actions of another party (...)
irrespective of the ability to monitor or control that other party}''~\cite{MayerRogerC.1995AIMO}. This is an apt definition in the context of AI systems where black-box systems are typical, and users typically do not possess the ability to assess, monitor, or enforce desirable properties of the system~\cite{eschenbach2021transparency}.
Consistent with this definition, \citet{ZhouJianlong2020Eopt} conceptualize trust in AI as the expectation that an AI agent will help achieve the user's goal. \citet{hoff2015trust} define trust as the tendency to take risk while believing there is a high chance of a positive outcome.

\begin{table}[b]
\centering
\caption{Elements of trust in AI}
\label{tab:trust-ai}
\vspace{-1em}
\renewcommand{\arraystretch}{0.8}
\footnotesize

\begin{tabular}{@{}lp{1.5cm}p{1.85cm}p{1.5cm}p{1.66cm}@{}}
\toprule
& \textbf{\citet{yang2022user}} 
& \textbf{\citet{afroogh2024trust}} 
& \textbf{\citet{nordheim2019initial}} 
& \textbf{\citet{seitz2022can}} \\
\midrule
\multirow{5}{*}{\rotatebox[origin=c]{90}{\parbox{1.25cm}{\centering Used in\\our study}}}
& Predictability &  & Predictability & Predictability \\
& Dependability & & & \\
& Alignment* & & & Alignment* \\
& & & Ease of use & Ease of use* \\
& & Transparency & & Transparency \\
\cmidrule{1-5}
\multirow{9}{*}{\rotatebox[origin=c]{90}{\parbox{2.5cm}{\centering Not used\\in our study}}}
& Ability & & & Ability \\
& & & Expertise & \\
& Faith & & & \\
& & & Reputation & \\
& & Reliability & \\
& & Accuracy & \\
& & & Risk & Risk \\
& Integrity & & & Integrity \\
& & Explainability & \\
& & & & Data Privacy \\
\bottomrule
\end{tabular}%

\caption*{\small Asterisk denotes concepts we refer to with an alternative name}
\vspace{-3em}
\end{table}

\subsubsection*{Elements of trust in AI}
To understand the properties and mechanisms through which trust in AI is formed, various decompositions of trust (in AI) have been proposed. In \tabref{tab:trust-ai}, we aligned four taxonomies of the elements of trust in AI. Clearly, different taxonomies overlap only partially, emphasizing different aspects of trust. \citet{yang2022user} and \citet{afroogh2024trust} conduct systematic literature reviews on user trust in AI sampling from various domains. \citet{nordheim2019initial} and \citet{seitz2022can} focus on trust in chatbots, a particularly interactive AI, conceptually close to collaboration, our topic of interest.
We rely on the subset of elements in these taxonomies that are applicable in our experiments. The definitions of those elements are given below. Based on these trust elements, in \secref{sec:questionnaire}, we assemble a questionnaire to elicit the trust attitude of the participants in our experiments towards AI.

\begin{description}[leftmargin=0.4cm, labelwidth=!, itemindent=0pt]
    \item[Ease of use] The simplicity of technology as perceived by users. This concept is labeled as ``usability'' in \cite{seitz2022can}.

    \item[Dependability] A system's consistency over time and across different situations~\cite{yang2022user}. This is in line with the definition of \citet{avizienis2004basic}: the ability of a system to avoid constant failure. 
    
    \item[Predictability] The consistency of the trustee's performance or behaviour in an extended period of time~\cite{yang2022user}.

    \item[Alignment] The degree of agreement between user- and AI intentions~\cite{yang2022user}. This is referred to as ``benevolence'' by \citet{yang2022user} and \citet{seitz2022can}, but we find ``alignment'' more consistent with recent results in human-AI topics~\cite{ji2025ai}.
   
    \item[Transparency] The degree to which users understand how AI functions and whether its decision-making process is clear~\cite{seitz2022can}.
    
\end{description}

\subsection{Human-AI Interaction}

\subsubsection*{Mixed-initiative interaction}
Mixed-initiative interaction characterizes human-AI interaction as a cooperative process in which both the human and agent contribute at appropriate times to facilitate effective teamwork~\cite{allen1999mixed}. Initiative is dynamic rather than fixed, meaning either the human or AI system may act as a leader, supporter, or independently based on the changing demands of the task. Throughout the interaction, roles are negotiated dynamically through a dialogue that determines when the AI system listens versus when it communicates.
Mixed-initiative AI systems mark a stark departure from passive AI components that wait for humans to initiate interaction, and instead, they proactively prompt humans.

\subsubsection*{Agentic AI} Agentic AI is a prime example of AI systems that can operate in a mixed-initiative mode.
Agentic AI characterizes autonomous systems that aim to imitate human behaviour to complete tasks with minimal human supervision~\cite{nisa2025agentic}. The semi-autonomous capabilities of these systems, in addition to reasoning, planning, context-aware interaction, and adaptability, allow these systems to perform complex, multitask problems. Rather than following predefined rules, agentic systems dynamically adapt to changing environments by contextualizing data in different formats, like audio, text, and images. These characteristics position agentic AI as tools that extend beyond automation and toward collaborative task execution, where agentic AI can act as semi-independent assistants to human workflows. Due to the broad scope of these systems, taxonomies categorize these systems based on their level of autonomy, their ability to learn, and their capabilities. These include reactive and proactive agents, limited memory agents, model-based agent, goal driven agents, theory of mind agents, and self-aware agents.

\subsection{Related work}
Closest to our work are studies that examine confidence and trust dynamics in collaborative human-AI settings.
\citet{li2025as} investigate the relationship between human self-confidence and AI confidence, finding that human self-confidence shifts to align with AI confidence during collaboration, and that this shift persists even after the interaction ends. An important difference compared to our work is that there is no measure of human \textit{confidence in AI} and that human self-confidence is related to specific problems of a task (e.g., how confidence are you in the label you assigned to this input?). In contrast, our work focuses on users' self-confidence independently of the task at hand (e.g., users are asked about their confidence after playing a game of Pac-Man, rather than being asked to assign their confidence to individual game choices they make).

\citet{park2026authorship} examine how writing with assistance from large language models (LLMs) affects users' self-efficacy and their trust in AI, and how these traits evolve over the course of the interaction. The authors find that LLM-assistance decreased self-efficacy but increased trust in AI.
\citet{sullivan2025transparency} study how varying levels of AI transparency influence users' trust and perceived reliability in AI systems. Their results indicate that higher levels of AI transparency are associated with higher reported trust and confidence in the system, increased perceived reliability, and improved ease of understanding. 
\citet{choung2023trust} explore whether trust influences the acceptance of AI technologies, finding that trust indirectly shapes users' willingness to adopt the technology by increasing perceived usefulness and positive attitudes toward AI.

\citet{ma2024you} explore how calibrating human self-confidence influences their confidence calibration, the alignment between an individual's reported confidence and their actual correctness. Their findings suggest that improving self-confidence calibration improves human-AI collaboration and leads to a more rational reliance on AI. These works establish that trust, confidence, and transparency play a key role in improving human-AI interactions. However, they largely examine these attributes without accounting for external pressures that may affect human-AI collaboration, such as time pressure (urgency) in our experiment.

Other works closest to ours are those that examine how urgency affects collaboration in settings between humans and automation. 
\citet{tatasciore2024can} examine how task time pressure and transparency affect the use of automated decision aids, finding that high time pressure decreases accuracy and increases perceived workload. While they note that increasing transparency improves trust and accuracy, it does not alleviate the negative effects of time pressure. 
Similarly, \citet{rieger2022human} investigate how time pressure influences human use of automated decision support systems (DSSs). The authors find that time pressure tends to reduce decision accuracy and that joint human-DSS performance is worse than automation-alone performance. 
Overall, prior research in human-AI collaboration tends to focus on confidence and trust in low-pressure settings or examines how urgency primarily affects accuracy. In contrast, our work directly investigates how task urgency influences human self-confidence and trust in AI systems, bridging the gap in the previously mentioned works.
\begin{figure*}[]
  \centering
  \includegraphics[width=0.8\textwidth]{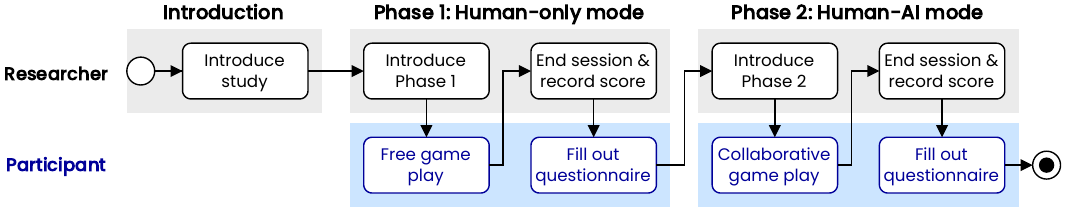}
    \caption{Experiment Overview}
    \label{fig:study-design}
\end{figure*}

\section{Study Design}\label{sec:study design}

To assess how human-AI interaction impacts trust in AI, we conducted an experiment with human participants, in which we compared participants' trust levels in AI before and after a human-AI interactive task.
We recruited 30 participants from various academic levels of computing and software programs, including senior undergraduates and graduate students (Master's, PhD). 

We conducted the experiments in person, in a controlled environment at McMaster University, Canada, between October 1 and November 1, 2025. Prior to our experiments, both the environment and the experimental tool were thoroughly tested to ensure reliability and consistency.
During each experiment, one participant and the lead researcher were present. We ensured that participants possess the required command in English to avoid language barriers.

\subsection{Experimental setup}\label{sec:experimental-setup}

\subsubsection{Experimental tool and task}
In our experiments, the participant plays a game of Pac-Man, a popular and well-known arcade game~\cite{rohlfshagen2018pacman}. The participant plays the game in two modes: first, without any AI agent involved; and subsequently, the control is passed to a reasonably trained AI agent that will ask for the participant's advice occasionally. The final score is the combined score of the human-only and human-AI interactive modes. To build emotional investment, the highest scoring player wins a book---a token of modest value, but sufficient to render the participant's stakes in the experiment high enough to test their trust in the AI agent under meaningful conditions.

We developed the Pac-Man game from the Pacman environment of Farama Foundation's Gymnasium framework\footnote{\url{https://gymnasium.farama.org/v0.28.0/environments/atari/pacman/}}, following established architectural principles of reinforcement learning environments~\cite{liu2026reference}. We trained the AI agent via reinforcement learning using the proximal policy optimization algorithm~\cite{schulman2017proximal} with the following hyperparameters: learning rate $\alpha=2.5e{-4}$, discount factor $\gamma=0.99$, episodes=100, rollout length $n\_steps=128$. This results in a reasonably trained AI agent that can play Pac-Man.

\subsubsection{Experiment overview}

As shown in \figref{fig:study-design}, each experiment consists of an introduction, and two experimental phases.

\begin{description}[leftmargin=0.4cm, labelwidth=!, itemindent=0pt]
   \item[Introduction] The lead researcher \textit{introduces the study} (5 minutes), and answers any questions from the participant. To build investment in the game, the participant is informed that the highest scoring participant wins a prize.
    
    \item [Phase 1: Human-only mode]
    The participant works \textit{without} the AI agent. After a brief \textit{Introduction of Phase 1}, the participant proceeds with a \textit{free game play} (5 minutes), trying to maximize their score. After five minutes of game play, the lead researcher ends the session and records the participant's score. To conclude Phase 1, and before working with the AI agent, the participant fills in a questionnaire about their general attitude towards AI agents. We use the results of this questionnaire as the baseline when we measure attitude towards AI \textit{after} the experiment. (For the details of the questionnaire, see \secref{sec:questionnaire}.
    
    \item [Phase 2: Human-AI mode] The participant works \textit{with} the AI agent. After a brief \textit{Introduction of Phase 2}, the AI agent takes over the control over the game. Occasionally (roughly, every 50 steps), the AI agent will prompt the participant for advice, rendering this step a \textit{collaborative game play}. The human participant responds by giving a directional advice using the directional buttons on the keyboard.

    \textbf{Different urgency modes.} To allow us to answer RQ2 (the effect of urgency in interaction on trust), the participant performs two rounds of interactive game play. In one round (5 minutes), the participant has unlimited time to provide the advice. In the other round (5 minutes), the participant has to provide the advice in five seconds, otherwise the agent will disregard it (urgency).

    \textbf{Counterbalanced within-subject design.}
    We use counterbalanced within-subject design~\cite{jhangiani2019research}, i.e, we expose participants to the two interaction modes in a different order. (Group 1: unlimited time interaction $\rightarrow$ limited-time interaction; Group 2: limited-time interaction $\rightarrow$ unlimited time interaction.)
    
    After ten minutes of game play, the researcher ends the session, records the participant's score, and the participant fills in the questionnaire (\secref{sec:questionnaire}).
    
\end{description}

\subsection{Questionnaire development}\label{sec:questionnaire}

We collect data about demographics and trust attitudes towards AI agents. For the latter, we use Likert-type rating scales to guide participants in expressing their answers. Likert-type scales are psychometric rating scales often employed in questionnaires to measure the attitude of participants towards a specific statement~\cite{joshi2015likert}.
Here, we measure the attitude of participants towards statements about trust and elements of trust in AI, as detailed in \secref{sec:trust-in-ai} and \tabref{tab:trust-ai}.

We design two questionnaires. The one filled in Phase 1 of the experiment (see \figref{fig:study-design}) elicits the participant's trust attitude towards AI \textit{in general}, before the interactive exercise. The one filled in Phase 2 elicits the participant's trust attitude specifically towards the AI agent in the experiment.

We use pairs of statements such as ``\textit{In general, I trust AI agents}'' in the first questionnaire and ``\textit{I trust this particular AI agent}'' in the second questionnaire. We ask similar pairs of questions about ease of use, dependability, predictability, alignment, and transparency---the applicable elements of trust extracted in \secref{sec:trust-in-ai} and \tabref{tab:trust-ai}. Our goal is to detect differences in the responses to these pairs before and after the intervention. To explain potential differences, in the second questionnaire, we elicit explanatory factors through Likert items such as ``\textit{Observing the AI agent in the experiment helped me understand the reasoning behind its decisions}'' (RQ1), and Likert items related to time pressure (RQ2).

We use a five-point Likert scale of \textit{1: Strongly disagree} to \textit{5: Strongly agree}, and analyze the data accordingly.\footnote{The complete questionnaire is available in the replication package.}

\subsection{Design trade-offs and threats to validity}

\subsubsection*{External validity} Our sample of experimental population may limit the generalization of results to a wider audience. To keep our experiment tractable, we recruited university students from computer science programs as participants. This population tends to be more experienced with AI tools than the general audience. However, this sample is still sufficient for our goal to measure \textit{change} in attitudes in response to the interventions in our study.

\subsubsection*{Internal validity}
Learning effect may give rise to internal validity.
To mitigate this threat, we use a counterbalanced within-subject~\cite{jhangiani2019research} design, i.e, expose participants to human-only and AI interaction modes in a different order and observe potential differences between the two groups.
Slight threats arise from the overloaded notion of ``\textit{trust}.'' To mitigate this threat, we provided participants with definitions of potentially ambiguous concepts.

\subsubsection*{Construct Validity}
In Likert-type data, a threat to construct validity stems from acquiescence bias, i.e., humans' tendency to agree with a statement in the questionnaire. Following questionnaire best practices, we mitigate this threat by using both positive and negative statements.
Another threat to construct validity is social desirability, i.e., answering a questionnaire in a way participants think will make them look good. We mitigate this threat by making the questionnaire fully anonymized.
Some threats to validity arise from the use of a Likert scales to assess perceived self-efficacy, instead of a better-suited psychometric scales, e.g., the generalized self-efficacy scale~\cite{schwarzer1995generalized}. Opting for Likert scales was an early design choice when we decided to measure multiple factors, not only self-efficacy.

\section{RQ1: Trust by human-AI interaction}\label{sec:results-rq1}
To validate our study setup, we first corroborate that interactions with an AI indeed increase trust in our experiment, in line with observations from the state of the art~\cite{wan2024it,wu2023human-in-the-loop,zhang2023trust,denno2024cognitive}.

\subsection{Improved trust in AI}\label{sec:rq1-trust}

\begin{figure}[h]
  \centering
  \begin{subfigure}{\linewidth}
    \includegraphics[trim = 3cm 15.5cm 0 0,clip,width=\linewidth]{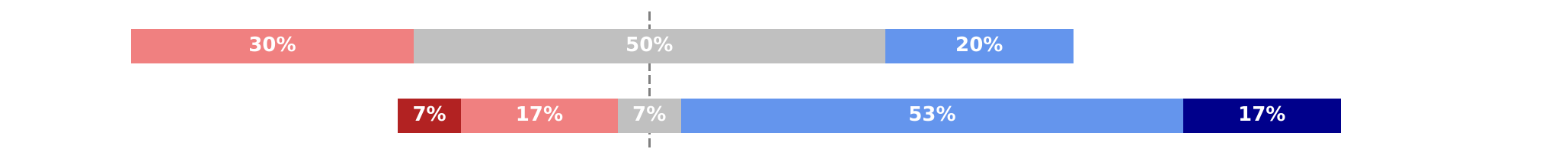}
    \caption{Trust in AI}
    \label{fig:likert-trust}
  \end{subfigure}\\[0.6em]
  \begin{subfigure}{\linewidth}
    \includegraphics[trim = 3cm 12.25cm 0 3.25cm,clip,width=\linewidth]{figures/results/likert/figure1.pdf}
    \caption{Ease of use}
    \label{fig:likert-easeofuse}
  \end{subfigure}\\[0.6em]
  \begin{subfigure}{\linewidth}
    \includegraphics[trim = 3cm 9.25cm 0 6.5cm,clip,width=\linewidth]{figures/results/likert/figure1.pdf}
    \caption{Dependability}
    \label{fig:likert-dependability}
  \end{subfigure}\\[0.6em]
  \begin{subfigure}{\linewidth}
    \includegraphics[trim = 3cm 6.25cm 0 9.5cm,clip,width=\linewidth]{figures/results/likert/figure1.pdf}
    \caption{Predictability}
    \label{fig:likert-predictability}
  \end{subfigure}\\[0.6em]
  \begin{subfigure}{\linewidth}
    \includegraphics[trim = 3cm 3.25cm 0 12.5cm,clip,width=\linewidth]{figures/results/likert/figure1.pdf}
    \caption{Alignment}
    \label{fig:likert-alignment}
  \end{subfigure}\\[0.6em]
  \begin{subfigure}{\linewidth}
    \includegraphics[trim = 3cm 0 0 15.5cm,clip,width=\linewidth]{figures/results/likert/figure1.pdf}
    \caption{Transparency}
    \label{fig:likert-transparency}
  \end{subfigure}\\
  \caption{\centering Participants' perception of trust in AI (\textit{a}) and its elements (\textit{b}--\textit{f}) before (top) and after (bottom) a human-AI interactive experience\\\footnotesize{\color{firebrick}{$\blacksquare$ Strongly disagree}~~\color{lightcoral}{$\blacksquare$ Disagree}~~\color{silver}{$\blacksquare$ Neutral}~~\color{cornflowerblue}{$\blacksquare$ Agree}~~\color{darkblue}{$\blacksquare$ Strongly agree}}}
  \label{fig:likert}
\end{figure}

\begin{figure}[t]
  \centering
  \includegraphics[width=0.9\columnwidth]{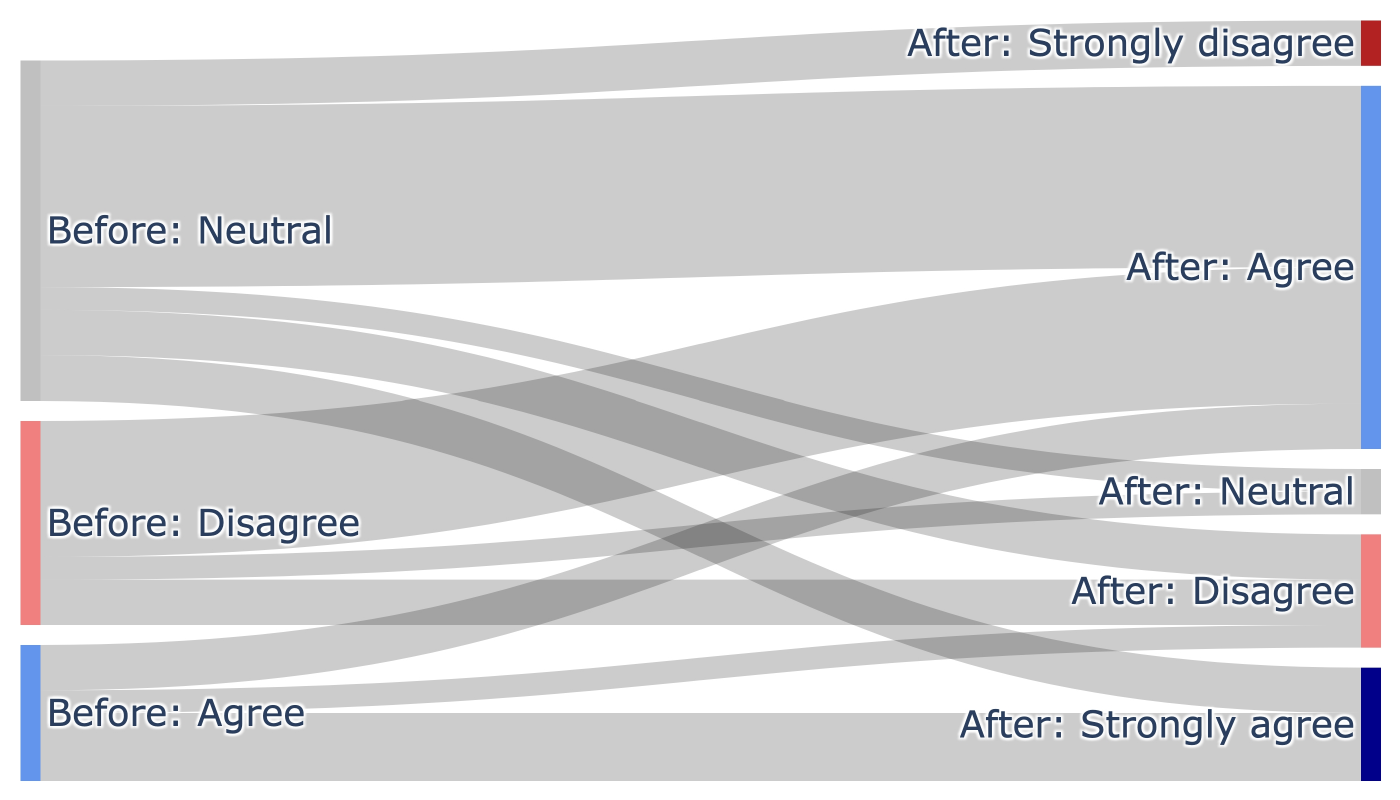}
  \caption{Changing trust attitudes of participants}
  \label{fig:sankey}
\end{figure}

\figref{fig:likert} reports the attitude responses before and after the experiment.
Shown in \figref{fig:likert-trust}, we measure a 50\% increase in trust as the previous number of agreements increases from 20\% to 70\% (strong agreement and agreement in total). At the same time, disagreement decreases only slightly, from 30\% to 24\%. A fraction of people, 7\%, remain neutral.
A more detailed look in \figref{fig:sankey} reveals that the 50\% post-intervention increase in trust is not solely due to previously neutral participants changing their attitude.

The eventual 70\% of participants who express trust in the AI agent comprises \xofy{5}{30}{participants} who were trustful before the intervention, \xofy{10}{30}{} previously neutral, and \xofy{6}{30}{} previously distrustful participants. Conversely, the previous 50\% neutral stratum becomes more trustful in \xofy{10}{30}{cases}, remains neutral in \xofy{1}{30}{cases}, and becomes distrustful in \xofy{4}{30}{cases}.\footnote{More details are available in the replication package.}

\keyobservation{Trust increases substantially, especially due to both neutral and distrustful participants developing trust.}

As shown in \figref{fig:likert-easeofuse}--\ref{fig:likert-transparency}, the known elements of trust do not change substantially, despite the clear increase in trust.
The perceptions of most factors slightly improves. The only exception being ease of use (\figref{fig:likert-easeofuse}), with a change of agreement from 90\% to 87\%.
The number of participants who agree that the AI agent is dependable (\figref{fig:likert-dependability}) increases by 6\%, and the number of those who disagree with this assertion decreases by about 13\%. A similar tendency can be observed in predictability (\figref{fig:likert-predictability}) with an increase in agreement by 10\% and a decrease in disagreement by 9\%; in alignment (\figref{fig:likert-alignment}) with an increase in agreement by 13\% and a decrease in disagreement by 4\%; and in transparency (\figref{fig:likert-transparency}) with an increase in agreement by 6\% and a decrease in disagreement by 10\%.

\keyobservation{Trust increases substantially, but known trust components remain essentially unaffected.}

\subsubsection*{Discussion} Our observations are in line with the established view in the state of the art, i.e., that trust increases with interaction. We did not detect change in the specific trust components that could pinpoint the key mechanism behind the improved trust attitude. One plausible explanation is that we did not select the right trust factors in our study design (\tabref{tab:trust-ai}). Another plausible explanation is that current trust taxonomies may not be able to describe causation between trust component and eventual trust. However, this investigation is not in the scope of our work; we merely aimed to establish that trust indeed increases in our setting.

\subsection{Preference against AI autonomy}

\begin{figure}[h]
  \centering
  \includegraphics[trim = 0 0 2cm 0,clip,width=\linewidth]{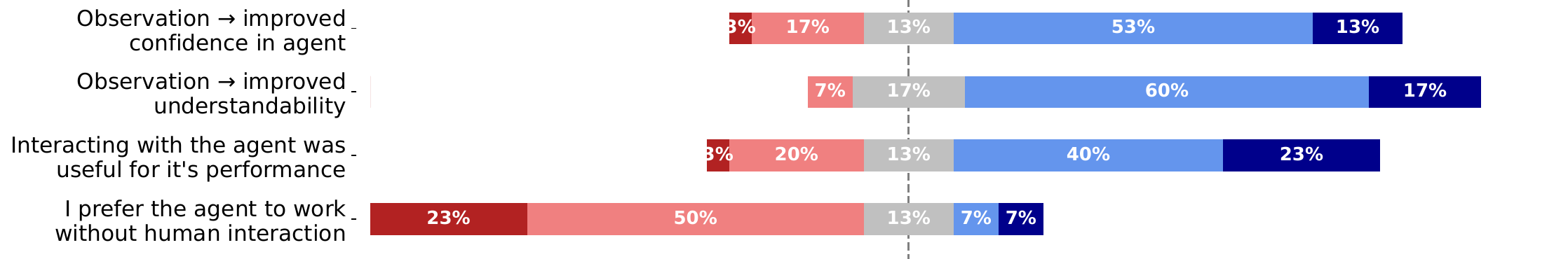}
  \caption{\centering Perception of interaction with the AI agent. (Likert items reasonably paraphrased for brevity.)\\\footnotesize{\color{firebrick}{$\blacksquare$ Strongly disagree}~~\color{lightcoral}{$\blacksquare$ Disagree}~~\color{silver}{$\blacksquare$ Neutral}~~\color{cornflowerblue}{$\blacksquare$ Agree}~~\color{darkblue}{$\blacksquare$ Strongly agree}}}
  \label{fig:collaboration}
\end{figure}

In the post-experiment questionnaire, we asked the participants for additional details, shown in \figref{fig:collaboration}, to be able to explain the mechanisms that effect trust. 

Following the interactive experience, \xofy{20}{30}{participants} agree that their confidence in the agent improved by observing its behavior; and only \xofy{6}{30}{} disagree with this assertion. \xofy{23}{30}{participants} agree that the understandability of the agent improved by observing its behavior; and only \xofy{2}{30}{} disagree.
These figures demonstrate that observation alone may improve key attitudes towards the AI (confidence in, and understandability of AI).
\xofy{20}{30}{participants} also agree that the agent's performance improved by the active human-AI interaction.

\xofy{23}{30}{participants} disagree with the suggestion that the AI agent should work on its own. Thus, despite the improved confidence in and understandability of the AI agent, participants prefer to remain part of the collaboration.
This observation excludes the possibility that humans' attitudes towards the AI agent changed positively due to the simplicity of the experimental task.

\keyobservation{Human-AI interaction is a valued mechanism, and humans prefer not to allow the AI agent to work on its own.}

\subsubsection*{Discussion} Despite the improved trust towards the AI, the participants are still reluctant to allow more autonomy to the AI and let it operate without human interaction. This is an unexpected development considering that AI's superior ability to play simple video games is well-documented and well-known~\cite{risi2020chess}. A plausible explanation is the lack of sufficient onboarding and therefore, the lack of willingness to give autonomy. Onboarding in AI software is often tied to the gradually increasing degree of autonomy the AI system is given, which increases trust calibration and usability compared to immediate full autonomy~\cite{kocielnik2019will}.
Another plausible explanation is that participants have pre-existing biases that our study was not designed to uncover.

\phantom{}

\begin{cframe}{RQ1: Improved trust by interaction}
    We corroborate that interactions with the AI agent indeed improve users' perceived trust in the agent. However, we remark that despite the increased trust, humans still prefer interactive human-AI collaboration rather than giving full autonomy to the AI.
\end{cframe}
\section{RQ2: The impact of urgency}\label{sec:results-rq2}

As explained in \secref{sec:experimental-setup}, in Phase 2, two groups of participants interacted with the AI differently. Participants in Group 1 first interacted with the AI agent without a time limit (i.e., no urgency) to provide advice, and subsequently, with time limit (i.e., urgency). Participants in Group 2 executed Phase 2 in the other way around, i.e., were exposed to urgency without prior experience with the AI.

\subsection{Significant difference in self-confidence}

\begin{figure}[h]
  \centering
  \begin{subfigure}{\linewidth}
    \includegraphics[trim = 0.8cm 0 3.7cm 1.25cm,clip,width=\linewidth]{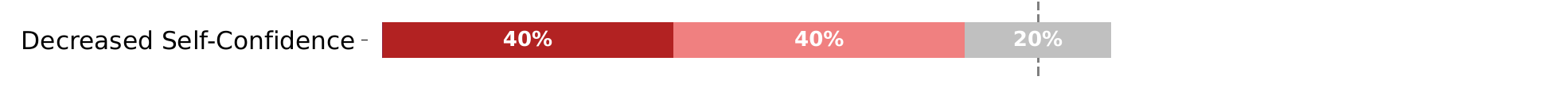}
    \caption{Group 1: unlimited $\rightarrow$ limited interaction time}
    \label{fig:likert-focus-confidence-group1}
  \end{subfigure}\\[0.5em]
  \begin{subfigure}{\linewidth}
    \includegraphics[trim = 0cm 0 4.5cm 1.25cm,clip,width=\linewidth]{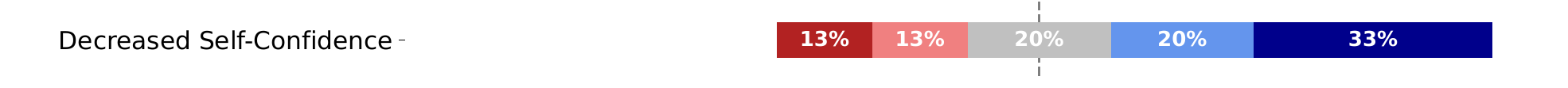}
    \caption{Group 2: limited $\rightarrow$ unlimited interaction time}
    \label{fig:likert-focus-confidence-group2}
  \end{subfigure}\\[0.5em]
  \begin{subfigure}{\linewidth}
    \includegraphics[trim = 0cm 0 4.6cm 1.25cm,clip,width=\linewidth]{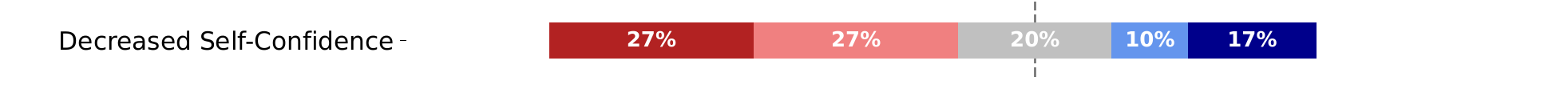}
    \caption{Total population}
    \label{fig:likert-focus-confidence-total}
  \end{subfigure}
  \caption{\centering Perceived effects of time pressure on self-confidence\\\footnotesize{\color{firebrick}{$\blacksquare$ Strongly disagree}~~\color{lightcoral}{$\blacksquare$ Disagree}~~\color{silver}{$\blacksquare$ Neutral}~~\color{cornflowerblue}{$\blacksquare$ Agree}~~\color{darkblue}{$\blacksquare$ Strongly agree}}}
  \label{fig:likert-focus-confidence-groups}
\end{figure}

The most important observation is shown in \figref{fig:likert-focus-confidence-groups}.

We observe substantial difference between the groups in terms of self-confidence. Most participants in Group 1, \xofy{12}{15}{} disagree or strongly disagree that their self-confidence decreased; and no one agrees with this statement. Conversely, in Group 2, only \xofy{4}{15}{participants} disagree and the majority of participants, \xofy{8}{15}{} agree that their self-confidence has decreased.

A chi-square test reveals that the difference between Group 1 and Group 2 is significant at $\alpha = 0.05$ with $p = 0.002$ ($\chi^2$ = 12.00), and the effect size (Cramér's V = 0.63) suggests strong association.

\keyobservation{Subjecting participants to time-pressured prompting by the AI has a statistically significant adverse impact on participants' self-confidence in our experiment.}

\subsubsection*{Discussion}

These figures suggest that subjecting participants to time-pressured prompting by the AI may decrease participant's self-confidence and that, at significant levels. However, this effect disappears when participants are given the opportunity to first interact with the AI without time constraints, i.e., being eased into the collaboration with the AI. Indeed, it seems that with reasonable time to get familiar with an AI tool, participants feel more confident about their own work.
As shown in \figref{fig:likert-focus-confidence-total}, this significant tendency of decreased self-confidence cannot be detected at the level of the total experimental population (i.e., by lumping the results from Group 1 and Group 2). It is, therefore, plausible that the difference between Group 1 and Group 2 in terms of self-confidence is due to the specific sequences of urgency modes.

\subsection{No difference in trust and in the preference against AI autonomy}

\begin{figure}[h]
  \centering
  \begin{subfigure}{\linewidth}
    \includegraphics[trim = 3cm 15.5cm 1cm 0,clip,width=\linewidth]{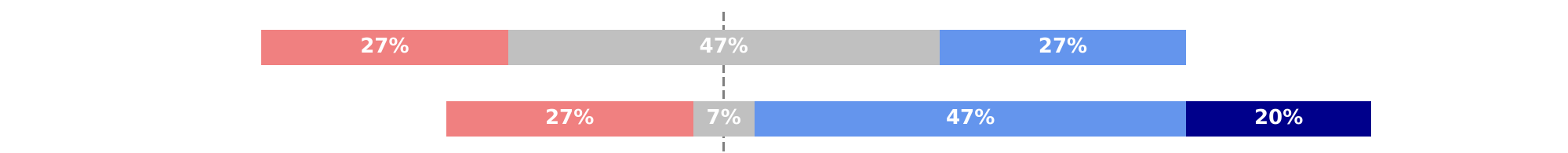}
    \caption{Group 1: unlimited $\rightarrow$ limited interaction time}
    \label{fig:likert-trust-group1}
  \end{subfigure}\\[0.5em]
  \begin{subfigure}{\linewidth}
    \includegraphics[trim = 2cm 15.5cm 2cm 0,clip,width=\linewidth]{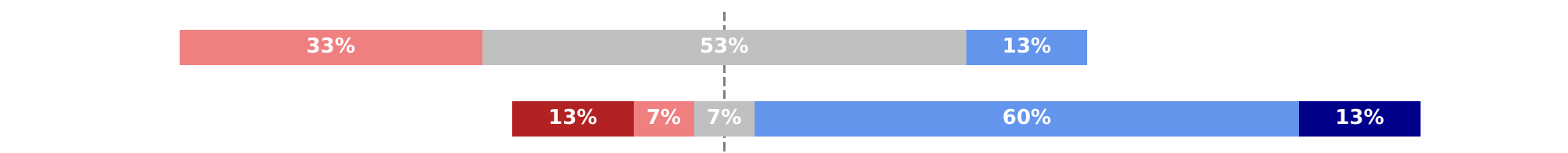}
    \caption{Group 2: limited $\rightarrow$ unlimited interaction time}
    \label{fig:likert-trust-group2}
  \end{subfigure}\\[0.5em]
  \begin{subfigure}{\linewidth}
    \includegraphics[trim = 1.4cm 15.5cm 1.6cm 0,clip,width=\linewidth]{figures/results/likert/figure1.pdf}
    \caption{Total population (as shown in \figref{fig:likert-trust})}
    \label{fig:likert-trust-total}
  \end{subfigure}
  \caption{\centering Attitude towards trust in the two groups before (top) and after (bottom) human-AI interaction\\\footnotesize{\color{firebrick}{$\blacksquare$ Strongly disagree}~~\color{lightcoral}{$\blacksquare$ Disagree}~~\color{silver}{$\blacksquare$ Neutral}~~\color{cornflowerblue}{$\blacksquare$ Agree}~~\color{darkblue}{$\blacksquare$ Strongly agree}}}
  \label{fig:likert-trust-groups}
\end{figure}

\figref{fig:likert-trust-groups} shows that trust attitude towards the AI changes in the same way in the two groups, and this change is consistent with the change observed in the total population (previously reported in \secref{sec:rq1-trust} and \figref{fig:likert-trust}).
Distrust decreases and trust increases, eventually reaching about the same level of trust in both groups (Group 1: \xofy{10}{15}{participants} agree or strongly agree; Group 2: \xofy{11}{15}{participants} agree or strongly agree).
This aligns with the trust attitude measured on the total population (\xofyp{21}{30}), shown in \figref{fig:likert-trust-total}.

\keyobservation{Urgency has no substantial effect on the eventual trust attitude. Trust increases similarly in both observed urgency modes.}

\begin{figure}[h]
  \centering
  \begin{subfigure}{\linewidth}
    \includegraphics[trim = 0 0.3cm 4.1cm 4.7cm,clip,width=\linewidth]{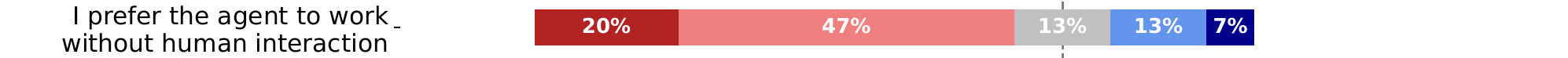}
    \caption{Group 1: unlimited $\rightarrow$ limited interaction time}
    \label{fig:likert-preference-group1}
  \end{subfigure}\\[0.5em]
  \begin{subfigure}{\linewidth}
    \includegraphics[trim = 0cm 0.3cm 5.3cm 4.7cm,clip,width=\linewidth]{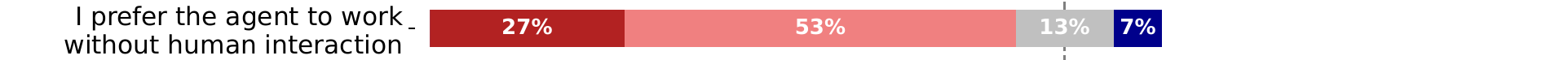}
    \caption{Group 2: limited $\rightarrow$ unlimited interaction time}
    \label{fig:likert-preference-group2}
  \end{subfigure}\\[0.5em]
  \begin{subfigure}{\linewidth}
    \includegraphics[trim = 0 0.3cm 5.5cm 4.7cm,clip,width=\linewidth]{figures/results/likert/label_figure2.pdf}
    \caption{Total population (as shown in \figref{fig:collaboration})}
    \label{fig:likert-preference-total}
  \end{subfigure}
  \caption{\centering Perceived effects of time pressure on users' interaction preference (Group 1 and 2) \\\footnotesize{\color{firebrick}{$\blacksquare$ Strongly disagree}~~\color{lightcoral}{$\blacksquare$ Disagree}~~\color{silver}{$\blacksquare$ Neutral}~~\color{cornflowerblue}{$\blacksquare$ Agree}~~\color{darkblue}{$\blacksquare$ Strongly agree}}}
  \label{fig:likert-interaction preference-groups}
\end{figure}

We do not observe a difference in participants' preferences for or against AI autonomy. As reported in \figref{fig:likert-interaction preference-groups}, the two groups show a similar tendency towards \textit{not} giving complete autonomy to the AI (Group 1: \xofy{10}{15}{participants} disagree or strongly disagree; Group 2: \xofy{12}{15}{participants} disagree or strongly disagree).
This aligns with the preference measured on the total population (\xofyp{22}{30}), shown in \figref{fig:likert-preference-total}.

\keyobservation{Urgency has no substantial effect on the participants' preference against giving complete autonomy to the AI.}

\subsubsection*{Discussion} The loss of self-confidence in our experiments manifested in response to unprepared urgency. That is, Group 2 was not allowed an unlimited-time joint work with the AI agent before time pressure has been applied. Group 1 has had the opportunity to work with the AI agent without being rushed. We conjecture that this interaction allowed Group 1 to understand the impact of their own contributions to the joint problem-solving process and value themselves accordingly.
Improved trust in the AI at the same time as loss of self-confidence is a cognitive asymmetry that may lead to cognitive dissonance which, in turn, may be detrimental to self-image, erode self-efficacy, and deteriorate self-esteem~\cite{stone2001self-standards}.

\phantom{}

\begin{cframe}{RQ2: The effects of (unprepared) urgency}
Our results indicate that urgency during human-AI interaction may have substantial (even significant) impact on human users' self-confidence when humans are not given sufficient time to get familiar with the AI. At the same time, even when self-confidence drops, trust in the AI is maintained.
\end{cframe}
\section{Discussion}\label{sec:discussion} 
We now discuss the results and elaborate on the key implications for software engineering researchers and practitioners.

\subsection{On the cognitive asymmetry of reduced self-confidence and improved trust}

One plausible explanation of the cognitive asymmetry between increased trust in the AI agent and decreased self-confidence is that introducing time pressure without proper training amplifies anxieties around AI's capabilities. Evidence from the state of the art shows that a common fear about AI---despite the obvious need for human intelligence and ingenuity in solving complex problems---stems from the reasoning capabilities of AI agents~\cite{diebel2025when}. Humans may (and often do) attribute anthropomorphic properties to AI agents~\cite{cheng2022human}, and may feel pressured by the abilities of this perceived virtual ``human'' that exceed actual human skills. This, in turn, may lead to a loss of confidence~\cite{alabed2022ai,chong2022human}. The increased focus we observe under time pressure (\figref{fig:likert-focus-confidence-groups}) may indicate a response to increased pressure in an attempt to keep pace with the AI. However, such cognitive pressure in long term is unlikely to be desirable as it is linked to increased burnout~\cite{schaufeli2014critical} and reduced productivity~\cite{amer2022occupational}.

There are numerous human and technical consequences of our observations.
It is easy to see how this cognitive asymmetry may lead to unsustainable human-AI collaboration on account of eroding self-efficacy, i.e., a person's belief in their ability to successfully organize and execute the actions required to achieve specific goals~\cite{bandura1977self-efficacy}.
Low self-esteem and reduced perceived competence are strongly associated with anxiety and stress in longitudinal studies~\cite{sowislo2013does}. Evidence from automated decision support systems also shows that time pressure tends to reduce decision accuracy~\cite{rieger2022human}, further impacting humans' perceived self-efficacy.
But anxiety and eroded self-esteem are not the only symptoms humans may develop in this setup. Humans may also develop overreliance on AI~\cite{ullrich2021development}. Although in our experiments, the participants overwhelmingly rejected the idea of allowing complete autonomy to the AI (\figref{fig:likert-interaction preference-groups}), it is possible that repeated exposure to new AI tools and problems could shift the participants' attitudes. Such experiments are outside the scope of our study and are left for future work.

In general, our observations and the subsequent mechanisms can be framed by the theory of cognitive dissonance, the psychological discomfort that arises from holding conflicting cognitions~\cite{stone2001self-standards}. Individuals have a motivation to seek consonance between their beliefs and actions~\cite{aronson1969theory}, and if consonance cannot be established, individuals tend to change their beliefs rather than their actions~\cite{festinger1957theory}, e.g., overestimating AI's indispensability or underestimating their own competence. It is plausible that unprepared urgency challenged the participants in our study in their belief that they are a useful party in the human-AI collaborative endeavor; to which the participants may have reacted by changing their belief, which manifested in the measured reduction in self-confidence.

From a technical point of view, these human implications may limit the automation level and adoption potential of AI. If AI tools stop being useful and empowering aids, their adoption will eventually slow down. It is, therefore, in every AI developer's and vendor's best interest to consider such barriers and actively help ease humans into human-AI settings.

\subsection{Self-efficacy as a software quality attribute}
To respond to the potential adverse implications of reduced self-confidence on self-efficacy---one's belief in their ability to successfully organize and execute the actions required to achieve specific goals~\cite{bandura1977self-efficacy}---we advocate for framing self-efficacy as a software quality attribute.
Self-efficacy influences how people approach challenges and persist in challenging (professional) situations, e.g., the ones that require AI assistance. This makes a good case for asserting explicit value to self-efficacy in the design of AI-enabled systems.

Framing social and individual sustainability properties as a quality attribute of software is not a new idea. Such directions have been thoroughly explored in the related body of knowledge, notably in the seminal works of \citet{lago2015framing}, \citet{abrahao2025software}, and \citet{naveed2024towards}. However, evidence shows that software engineering has limitations in embracing individual and social sustainability properties,~\cite{gustavsson2020blinded,shajari2025bridging}. To address this shortcoming, individual and social sustainability properties and particularly, self-efficacy ought to be treated as a quality attribute that can be systematically assessed, optimized, and engineered into software systems.

Software engineering as a profession itself would benefit from such improvements as well. Self-efficacy in software engineering has been associated with satisfaction, performance, and engagement with work and their teams among software engineers in industry~\cite{power2024self-efficacy}. Software engineers with high self-efficacy exhibit more proactive, socially engaged, and confident behavior in development contexts~\cite{ribeiro2023understanding}. Already at the early stage of engineering education, self-efficacy is a strong predictive marker of academic performance, motivation, persistence, and skill development~\cite{power2024self-efficacy}.
Therefore, our recommendation to treat self-efficacy as a quality of software applies to software used by software engineers, too, including IDEs, design tools, and documentation assistants. Modern IDEs already provide generative AI features, such as copilots, putting software engineers in a particularly susceptible position.

Promoting self-efficacy to a software quality attribute requires methodological and tool support, as well as thorough empirical evaluations. We recommend software engineering researchers to expand software evaluation methods by validated psychometric tools, standardized metrics, and experimental protocols, preferably of longitudinal nature to detect long-term tendencies of attitude change. Human factors research on trust and automation~\cite{hoff2015trust} provides an established foundation for incorporating self-efficacy--related measures into software engineering experiments. Research in human-computer interaction (HCI) and decision support indicates that transparent communication of uncertainty can improve calibrated trust and decision quality (e.g., visual and probabilistic uncertainty representations~\cite{kay2016when}; explanation and rationale-sharing~\cite{amershi2019guidelines}; and adaptive confidence displays~\cite{kocielnik2019will}).

\subsection{Organizational dimensions: socio-techno-economic risks and sustainable adoption}
The implications of our observations extend beyond individuals' cognition and may impact the broader socio-techno-econmic context in which the investigated human-AI interactions are typical. Widespread digital transformation triggered rapid adoption of AI systems in organizations~\cite{squicciarini2021demand} with the expectation that AI will improve competitive advantage through increased productivity and problem-solving capabilities. However, if AI systems are not introduced and aligned with business processes properly, organizations may expose their employees to the unwanted effects outlined previously: lower self-confidence, decreased self-efficacy, less satisfaction from work, and anxiety.
Time-pressure may also incentivize superficial decision-making~\cite{seitz2022can}, over-reliance on automation~\cite{tatasciore2024can}, or avoidance of responsibility~\cite{elzein2019shared}. These are just some of the socio-economic risks organizations face when engaging with AI systems.

Both from a social responsibility and economic point of view, organizations should feel motivated to develop mechanisms to mitigate these risks.
Gradual onboarding and its variants---e.g., scaffolded introduction, progressive training---allow users to get familiar with the features and working modes of the system. Gradual onboarding is often tied to the increasing degree of autonomy an (AI) system is given, which increases trust calibration and usability compared to immediate full autonomy~\cite{kocielnik2019will}. Our results, specifically in \figref{fig:likert-interaction preference-groups} show that full autonomy of AI is indeed not desired by the participants, even in a simple task as an arcade game (given that sufficient investment his facilitated). In other forms of onboarding, features are introduced gradually to users~\cite{amershi2019software}. Such strategies could be used, e.g., to allow users to get comfortable with an agentic AI by restricting the AI's proactive prompting features, i.e., restricting it to a passive component the user controls. This removes urgency from the human-AI setting and invokes the mechanism we observed in experimental Group 1, i.e., self-confidence may be preserved. As demonstrated, onboarding is not a mere training activity but a complex deployment strategy organizations need to align with their digital transformation approach and existing business processes.

In companies with lower digital adeptness, upskilling plays an important role in establishing elementary working primitives that enable joint work between humans and AI~\cite{ccallari2025meaningful}. Targeted training has also been shown to improve technology acceptance~\cite{molino2020promotion}, reduce technology misuse~\cite{marler2006training}, and increase performance through improved self-efficacy~
\cite{torkzadeh1999computer}. Specifically in the case of AI systems in a work environment, upskilling can help employees adopt accurate mental models of system capabilities and properly calibrate their self-confidence~\cite{khot2024technology-enabled}. Organizations should be motivated to implement upskilling programs as such endeavors make organizations more resilient to future change---an added benefit that positively impacts the employees as well. Similar to onboarding, upskilling is not a simple training activity either, and must be situated in the complex socio-techno-economic landscape of organizations.

At the end of the day, sustainable adoption of AI systems demands that organizations consider the complex interplay of social, technical, and economic factors. In this work, specifically, we call for proper onboarding and upskilling mechanisms to retain human values and to facilitate employee satisfaction in an increasingly digitalized organization. Alas, social factors are still often overlooked as a sustainability factor~\cite{gustavsson2020blinded}. This is apparent, e.g., in top-down strategic programs such as the Twin Transition, framed in the European Green Deal, that has a demonstrated blind spot for social and individual aspects in sustainable digital transformation~\cite{shajari2025bridging}.

We recommend researchers to investigate AI deployment strategies that regularly evaluate social and individual aspects in organizations, and align well with the organization's economic goals. Studying organizational adoption patterns and using structured frameworks, such as SusAF~\cite{basmer2021susaf}
\section{Conclusion}\label{sec:conclusion}

In this paper, we reported on our empirical study on how urgency in human-AI interaction may lead to reduced self-confidence. At the same time, trust in the AI agent typically still improves in humans. We draw the attention to this cognitive asymmetry because it has numerous potentially adverse implications on humans who interact with AI systems, e.g., reduced self-efficacy and performance. Given that the latest wave of mixed-initiative AI systems---e.g., agentic AI---often actively prompt humans, such urged interactions may appear at an increasing rate.
We observe, however, that humans who are eased into a human-AI collaborative setting may overcome this problem and retain self-confidence. Thus, we recommend AI adopters (teams, organizations, companies) to consider upskilling and training before exposing humans to time-constrained situation.

We recommend researchers to investigate interaction mechanisms that aid the retention of self-confidence in human-AI settings, such as confidence calibration, and regular evaluation of psychometric properties. We advocate for promoting self-efficacy to a software quality metric to enable the systematic engineering of human-centered AI software, and we urge the development of supporting design methods, design principles, and UX patterns.
\subsubsection*{Acknowledgment.}
We acknowledge the support of the Natural Sciences and Engineering Research Council of Canada (NSERC), DGECR2024-00293 (End-to-end Sustainable Systems Engineering).

\bibliographystyle{ACM-Reference-Format}
\bibliography{bib/references}

\end{document}